\documentclass[a4paper,twoside]{article}

\usepackage{epsfig}
\usepackage{subcaption}
\usepackage{calc}
\usepackage{amssymb}
\usepackage{amstext}
\usepackage{amsmath}
\usepackage{amsthm}
\usepackage{multicol}
\usepackage{pslatex}
\usepackage{apalike}
\usepackage{algorithm2e}
\usepackage[bottom]{footmisc}
\usepackage{xurl} 
\usepackage{multirow}

\usepackage{SCITEPRESS}     

\begin{document}

\title{Billboard in Focus: Estimating Driver Gaze Duration from a Single Image}

\author{\authorname{Carlos Pizarroso\sup{1}\orcidAuthor{0009-0002-0268-199X}, 
Zuzana Berger Haladov\'a\sup{1}\orcidAuthor{0000-0002-5947-8063}, Zuzana \v{C}ernekov\'a\sup{1}\orcidAuthor{0000-0002-7617-4192} and Viktor Kocur\sup{1}\orcidAuthor{0000-0001-8752-2685}}
\affiliation{\sup{1}Faculty of Mathematics, Physics and Informatics, Comenius University Bratislava, Slovakia}
\email{\{carlos.pizarroso, haladova, cernekova, kocur\}@fmph.uniba.sk}
}


\abstract{Roadside billboards represent a central element of outdoor advertising, yet their presence may contribute to driver distraction and accident risk. This study introduces a fully automated pipeline for billboard detection and driver gaze duration estimation, aiming to evaluate billboard relevance without reliance on manual annotations or eye-tracking devices. Our pipeline operates in two stages: (1) a YOLO-based object detection model trained on Mapillary Vistas and fine-tuned on BillboardLamac images achieved 94\% mAP\@50 in the billboard detection task (2) a classifier based on the detected bounding box positions and DINOv2 features. The proposed pipeline enables estimation of billboard driver gaze duration from individual frames. We show that our method is able to achieve 68.1\% accuracy on BillboardLamac when considering individual frames. These results are further validated using images collected from Google Street View.}

\onecolumn \maketitle \normalsize \setcounter{footnote}{0} \vfill

\section{\uppercase{Introduction}}
\label{sec:introduction}

Roadside billboards are one of the most popular forms of outdoor advertising, designed to capture the attention of both pedestrians and drivers. However, while this method is effective for marketing purposes, the visual stimuli from these advertisements can become a significant source of distraction, potentially increasing the risks of traffic accidents \cite{OVIEDOTRESPALACIOS201985} \cite{harasimczuk_maliszewski_olejniczak-serowiec_tarnowski_2018}. Moreover, the overabundance of visual elements in urban environments contributes to what is often described as visual pollution, which can interfere with drivers’ awareness of the environment and decision-making \cite{land13070994}. Therefore, understanding how drivers visually interact with billboards is essential for improving road safety and managing the visual complexity of urban spaces. \\


The driving task is characteristically complex and attention-intensive, requiring constant monitoring of the environment, anticipating dangers, and precise motor coordination. Consequently, any visual distraction, especially those engineered to attract gaze, can pose a serious threat to safety. Existing studies highlight that glances towards advertisements can last up to one second or longer, well beyond what is considered safe, especially when leading to animated or high-luminance digital billboards \cite{COSTA2019127} \cite{BROME2021226} \cite{CRUNDALL2006671} \cite{ZALESINSKA2018439}. Yet, most prior analyses rely on manual annotations or eye-tracking setups that are costly, time-consuming, and difficult to scale. \\


Building on the earlier work that introduced the BillboardLamac dataset \cite{10308914}, this study presents a two-stage pipeline combining two modules (1) a YOLO object detector \cite{ultralytics2023yolo} trained on Mapillary Vistas \cite{Neuhold_2017_ICCV} and fine-tuned using the BillboardLamac dataset, and (2) a classification model that estimates the duration of driver gaze toward billboards into predefined attention levels utilizing DINOv2 features \cite{oquab2023dinov2} in combination with the detected bounding boxes. Unlike the previous approach, where the driver gaze duration toward billboards was estimated using object tracking, fixation preprocessing using multiple videos of drivers wearing eye-tracking glasses, our system detects billboards and gaze-related attention levels from single images. This removes the need for multi-frame tracking or video post-processing and even allows the pipeline to operate on individual images. By combining a YOLO-based detector with a gaze duration classifier, the proposed method offers a more scalable, accessible, and deployment-ready framework for studying how roadside advertisements relate to driver visual behavior.

We evaluate our method on the BillboardLamac dataset. Compared to the original method introduced by \cite{10308914}, we achieve lower accuracy 69\% vs. 75.9\% when considering aggregated decisions for individual billboards. However, this decrease comes with the significant advantage of the possibility of applying our method on single images where our best model reaches 68.1\% accuracy.

We further validate our method by collecting a dataset of Google Street View images of billboards from the BillboardLamac test set. Using this dataset, our best method achieves 66.3\% accuracy, showing the potential for future applications of billboard gaze duration estimation.

We make our code and data publicly available.\footnote{\url{https://github.com/carlos-p-t/billboard-in-focus}}

\section{\uppercase{Related Work}}

Driver distraction caused by roadside advertisements has been widely studied in behavioral and traffic safety research, with multiple studies demonstrating that billboards can capture gaze and delay driver reactions \cite{OVIEDOTRESPALACIOS201985} \cite{refId0}. However, most of this 
research relied on controlled experiments or simulation-based studies, but advances in computer vision and eye tracking have allowed real-world analyses of driver attention. A key contribution to this domain is the study that led to the creation of the BillboardLamac dataset \cite{10308914}, which introduced a multi-stage pipeline combining object detection, gaze tracking, and machine learning to assess billboard saliency. The dataset contains over 155,000 annotated billboard instances and fixation data, providing a strong foundation for automated attention analysis.

\subsection{Object Detection for Billboard Analysis}

Deep learning-based object detectors have become increasingly popular for analyzing roadside scenes and localizing advertisements. Two main types of detectors are commonly used: two-stage models, such as Faster R-CNN \cite{8482613}, which generate region proposals and then refine classification; and one-stage models, such as YOLO \cite{ultralytics2023yolo}
which perform both steps in a single pass to achieve real-time performance. 
Comparative studies \cite{9666654} \cite{s20164587} and recent work on billboard salience estimation \cite{Berger_Haladova_2024} show that YOLOv5 achieves performance comparable to or exceeding that of Faster R-CNN and SSD \cite{liu2016ssd}.
Another approach to object detection leverages Transformer-based networks \cite{carion2020end}, which model long-range dependencies using self-attention mechanisms \cite{vaswani2017attention}. 


\subsection{Datasets for Billboard and Scene Understanding}

Successful deep learning models rely on diverse and well-annotated training data. Mapillary Vistas \cite{Neuhold_2017_ICCV} provides large-scale, street-level imagery for general scene understanding, serving as a base for pretraining detection models. Conversely, BillboardLamac \cite{10308914} provides fine-grained billboard annotations and associated driver gaze data, enabling the exploration of visual saliency and attention mechanisms in real-world contexts. Together, these datasets support both broad scene understanding and focused attention modeling in driving contexts.



\subsection{Automation in Gaze and Attention Estimation}

Recent developments in gaze inference have focused on automating gaze–object association to reduce the reliance on manual eye-tracking annotation. For example, object-based gaze inference \cite{hou2022traffic} estimates likely gaze targets directly from visual context, mitigating subjectivity in labeling and enabling scalable attention analysis. Such approaches demonstrate the growing potential of integrating visual perception modeling with automated scene understanding. \\

Building upon previous studies linking gaze behavior and billboard saliency \cite{10308914}, the present work extends this research direction by integrating deep learning-based billboard detection and saliency classification into a unified, fully automated pipeline for assessing billboard relevance and driver distraction in real-world driving imagery.

\section{\uppercase{Data}}

\label{sec:data}

Two datasets were used to train and evaluate our proposed pipeline: a subset of the Mapillary Vistas \cite{Neuhold_2017_ICCV} and BillboardLamac \cite{10308914}. These datasets were selected for their complementary characteristics. Mapillary Vistas provides a large-scale urban imagery supporting billboard detection pre-training, while BillboardLamac offers a focused and gaze-annotated collection for fine-tuning and attention analysis. \\


\textbf{Mapillary Vistas} is a large-scale, street-level dataset widely used for semantic segmentation and object detection tasks. It contains more than 25,000 manually annotated high-resolution images from diverse urban and rural scenes. Since billboard instances are included as a labeled object class, a subset of 9,700 images was selected for this study due to computational constraints (6,000 for training, 1,200 for validation, 2,500 for testing). The subset was balanced to include both billboard and non-billboard scenes, ensuring adequate exposure to negative samples during training.\\

\textbf{BillboardLamac} dataset was specifically developed to support research on roadside advertisement detection and driver gaze analysis. It contains high-resolution images extracted from eight driving sessions recorded using Tobii Glasses~3 eye-tracking technology, worn by eight drivers who followed an identical route around Lamač, Bratislava. Each session was recorded under real traffic conditions along the same urban path to ensure the consistency of the environment. In total, 145 unique billboards were identified and labeled with three driver gaze duration classes: \textit{none}, \textit{medium},  and \textit{long}, corresponding to increasing fixation time from no fixation, to fixation up to 250 ms, and fixation above 250 ms. The complete dataset comprises 155{,}912 labeled frames suitable for gaze-based classification. 

The complete BillboardLamac dataset, including driving videos, annotated frames, and metadata, is publicly available on Hugging Face.\footnote{
\url{https://hf.co/datasets/carlospizarroso/BillboardLamac}
} The dataset supports both billboard detection and driver gaze duration estimation tasks and includes detailed documentation and precomputed spatial features.

To allow a fair comparison, \cite{10308914} suggest a train/test split with the test containing frames for 29 of the total of 145 unique billboards. We use this split to evaluate classification accuracy.

\section{\uppercase{Proposed Method}}


In this section we introduce our proposed method for driver gaze duration estimation on the level of individual frames. In contrast to previous work \cite{10308914}, our proposed method does not require aggregation of information across multiple frames from multiple videos, thus enabling automatic estimation of billboard visibility and driver attention without requiring eye-tracking devices, providing a scalable solution for real-world driving imagery.

Our method has two stages: billboard detection using the YOLO object detection framework \cite{ultralytics2023yolo} and classification via the FLAML framework \cite{wang2021flaml} relying on the detected bounding box position and visual features extracted using DINOv2 \cite{oquab2023dinov2}. 

In this section we first describe the object detector and its results, then we describe the classifier. The results for classification, including ablation studies are provided in Section~\ref{sec:results}.


\subsection{Billboard Detection}

The object detector relies on YOLO-based architectures trained in two stages: pre-training the subset of Mapillary Vistas dataset, followed by fine-tuning on the BillboardLamac dataset (see Sec.~\ref{sec:data}). The datasets were not merged, allowing the model to first learn general street-scene representations and then adapt specifically to Billboard features. For fine-tuning on the BillboardLamac dataset we used 6000 training images, 1200 validation images and 2500 images for testing. We trained two models YOLOv8 and YOLOv11 \cite{ultralytics2023yolo}.
Both networks were trained for 50 epochs using hyperparameters optimized on the validation set. The two best-performing models were selected according to mAP@50 and Recall.

\subsubsection{Billboard Detection Results}

\begin{table}[h]
\centering
\renewcommand{\arraystretch}{1.1}
\begin{tabular}{|l|c|c|}
\hline
\textbf{Metric} & \textbf{YOLOv8} & \textbf{YOLO11} \\ \hline
Precision & 45.4 & 51.0 \\ \hline
Recall & 50.8 & 48.0 \\ \hline
mAP@50 & 45.5 & \textbf{46.7} \\ \hline
mAP@50--95 & 32.3 & 34.7 \\ \hline
\end{tabular}
\caption{\textit{Base Training Results}: Detection performance of YOLOv8 and YOLO11 on Mapillary Vistas test set.}
\label{tab:detmapillary}
\end{table}

Among multiple YOLOv8 and YOLO11 models trained on a subset of the Mapillary Vistas dataset, the best configurations were selected based on mAP@50 and Recall. Table~\ref{tab:detmapillary} reports the metrics for these models on the Mapillary Vistas test set, showing that YOLO11 achieves higher precision and recall compared to YOLOv8.

\begin{figure}[ht]
  \centering
   {\epsfig{file = 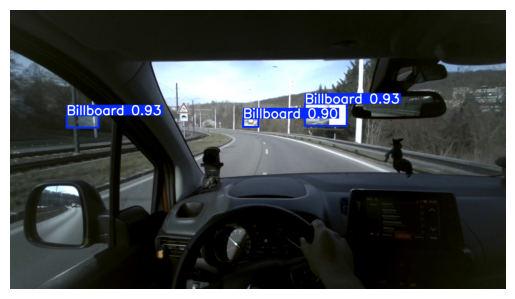, width = 7.5cm}}
  \caption{Example of billboard detections using the fine-tuned YOLO11 model.}
  \label{fig:detexample}
 \end{figure}

\begin{table}[h]
\centering
\renewcommand{\arraystretch}{1.1}
\begin{tabular}{|l|c|c|}
\hline
\textbf{Metric} & \textbf{YOLOv8} & \textbf{YOLO11} \\ \hline
Precision & 89.2 & 89.8 \\ \hline
Recall & 91.1 & 91.9 \\ \hline
mAP@50 & 93.8 & \textbf{94.0} \\ \hline
mAP@50--95 & 77.1 & 79.4 \\ \hline
\end{tabular}
\caption{\textit{Fine-Tuning Results}: Detection performance of YOLOv8 and YOLO11 on BillboardLamac test set.}
\label{tab:finetuning}
\end{table}

After fine-tuning both models on the BillboardLamac dataset, performance improved substantially across all metrics. As shown in Table~\ref{tab:finetuning}, the YOLO11-based detector achieved the highest performance, with 0.940 mAP@50 and 0.919 Recall, confirming that the model effectively adapts from general street-scene features to the specific visual characteristics of billboards. Qualitatively, the fine-tuned model correctly identified billboards under varying illumination, size, and occlusion conditions (Fig.~\ref{fig:detexample}). For further experiments we use the trained YOLO11 model.

\subsection{Gaze Duration Estimation}

In \cite{10308914}, the authors classify each billboard using features aggregated from multiple videos into three classes based on driver gaze duration. Likewise, we attempt to classify each detected billboard into the same classes, but instead of aggregation over full videos we consider a setup where only one frame is available. In order to do this, we intend to rely on the detected position of the bounding box and visual features extracted from the images.

Our initial experiments with deep convolutional neural networks for classification yielded weak results mainly due to a lack of diversity in the training data (only 145 billboard instances in the full dataset) and overfitting. Therefore, we opted to extract the DINOv2 features \cite{oquab2023dinov2} and use them in combination with the positions of the detected bounding boxes within an automated machine learning framework FLAML \cite{wang2021flaml}. 

For training we use the same split as \cite{10308914}. We obtain billboard detections using our trained YOLOv11 object detection network described in the previous subsection. For training, we keep only the 10 frames where the billboard bounding box was the largest for each billboard instance and driver, leaving us with 8,257 detections (for some billboards there were fewer than 10 detections). For all such detections, we extracted the 384-dimensional feature vector corresponding to the \textit{CLS} token of DINOv2-Small model. We also extracted the same features for the cropped images of the detected billboards. To reduce the dimensionality, of these features, we obtain a transformation using PCA on the features extracted from the training data. We denote these features as \textit{I$_{full}$} for full images and \textit{I$_{crop}$} for cropped images. To represent the bounding boxes as features, we use a four-dimensional vector with the x and y coordinates of their centers and width and height normalized to the image dimensions. We denote these features as \textit{B}.

For hyperparameter optimization we use k-fold cross-validation with splits based on billboard instance IDs. We train an ensemble of models using all base classifiers provided by the FLAML framework. Based on validation results, for PCA-based features we keep only the 3 most significant components. This reduction in dimensions significantly helps reduce overfitting. Using this strategy, we evaluate several combinations of features. The results and comparison to the previous method is provided in the next section.


\section{\uppercase{Results}}
\label{sec:results}

\begin{table*}
\center
\begin{tabular}{cccccccccc}

\multicolumn{4}{c}{Features} & \multicolumn{3}{c}{Per-detection} & \multicolumn{3}{c}{Aggregated} \\ \hline

\textit{Orig.} & \textit{B} & \textit{I$_{full}$} & \textit{I$_{crop}$} & Accuracy & Macro F1 & Micro F1 & Accuracy & Macro F1 & Micro F1 \\ \hline

 \checkmark & & & & - & - & - & \textbf{75.9} & \textbf{65.9} & \textbf{73.8} \\ 





 \checkmark & \checkmark & \checkmark & \checkmark & - & - & - & \textbf{75.9} & \textbf{65.9} & \textbf{73.8} \\

\hline








 &\checkmark& & &62.2&39.1&56.1&62.1&36.9&52.3\\
 & &\checkmark& &63.0&26.3&49.0&55.2&23.7&39.2\\
 & & &\checkmark&63.2&25.9&49.0&55.2&23.7&39.2\\
  & &\checkmark&\checkmark&63.2&27.7&50.2&55.2&23.7&39.2\\
 &\checkmark&\checkmark& &\textbf{68.1}&47.4&\textbf{62.9}&69.0&47.9&63.1\\
 &\checkmark& &\checkmark&63.8&36.4&55.5&58.6&30.9&46.3\\
 &\checkmark&\checkmark&\checkmark&67.3&\textbf{48.3}&62.5&69.0&46.5&62.2\\

\end{tabular}
\caption{The results on the test set of the BillboardLamac dataset. We compare how usage of different features affect the results: \textit{Orig.} - features aggregated over multiple videos as used by \cite{10308914}, \textit{B} - position of the detected billboard bounding box, \textit{I$_{full}$} - DINOv2 features for the full image, \textit{I$_{crop}$} - DINOv2 features for the cropped image.} 
\label{tab:results}
\end{table*}

\begin{figure}[h]
  \centering
   {\epsfig{file = 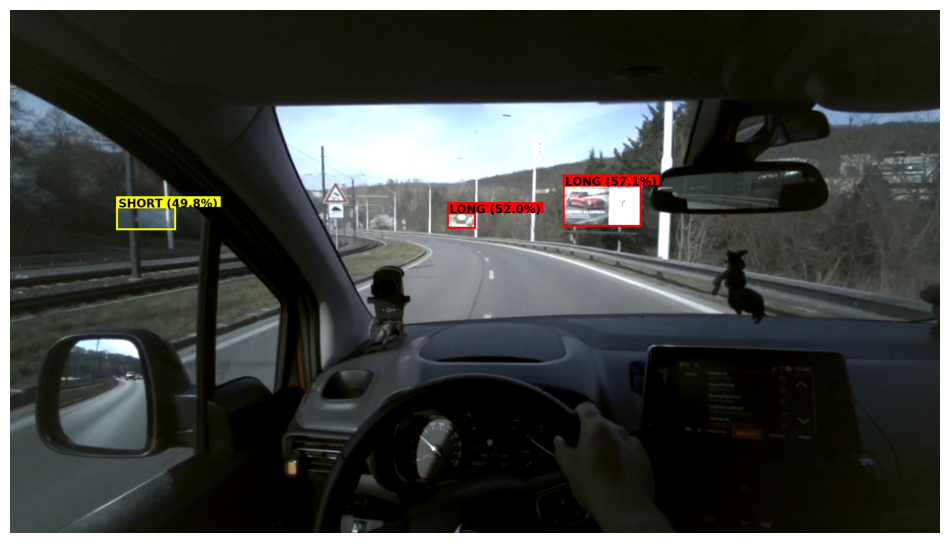, width = 7.5cm}}
  \caption{Example of classified billboards into gaze duration categories.}
  \label{fig:classexample}
 \end{figure}

\begin{figure}[h]
\centering
\epsfig{file=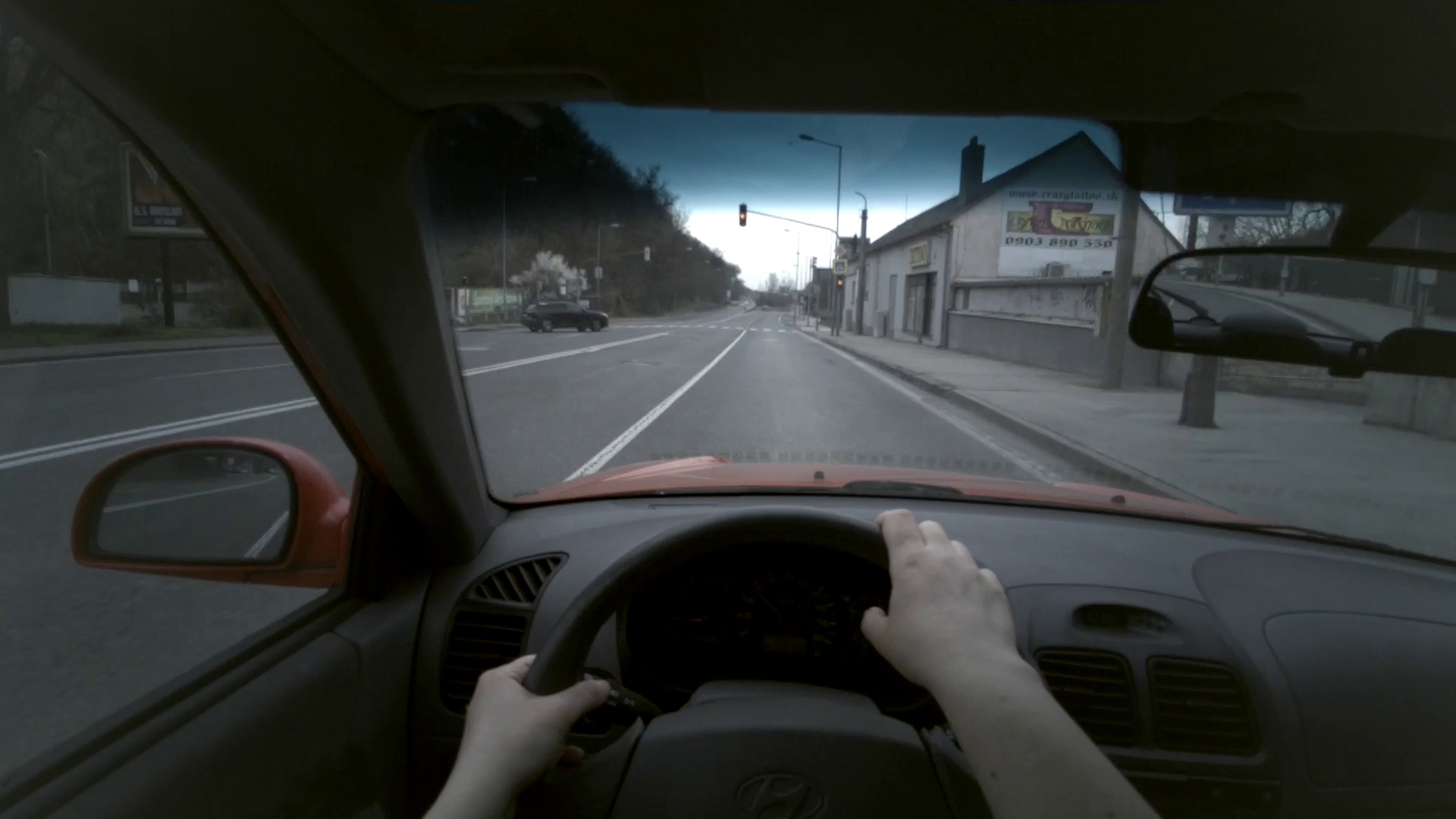, width=7.5cm}

\vspace{1ex}

\epsfig{file=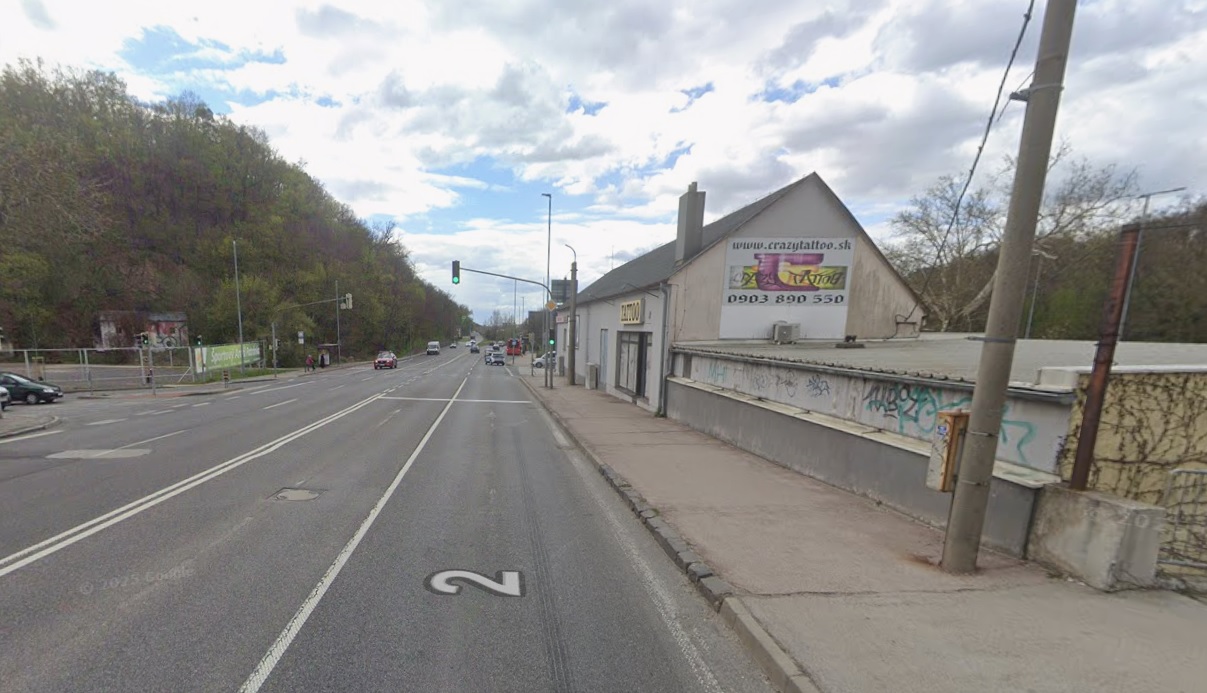, width=7.5cm}

\vspace{1ex}

\epsfig{file=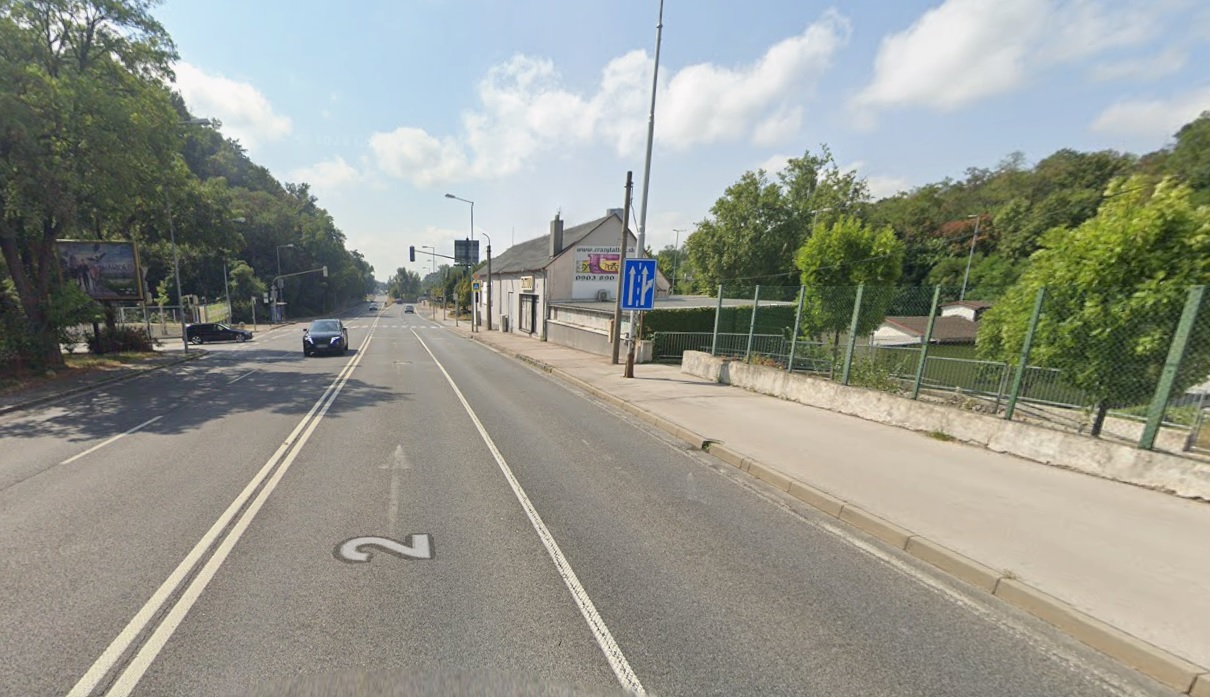, width=7.5cm}

\caption{Billboard from the BillboardLamac test set (top) and its corresponding Google Street View example in different views (bottom).}
\label{fig:gsv}
\end{figure}

In this section, we provide the evaluation of the proposed method. To allow for comparison with the method by \cite{10308914}, we assume two scenarios: classification per detection and aggregated classification. For the per-detection classification, we simply provide the class prediction for each detected billboard (see Figure~\ref{fig:classexample}). For evaluation, we considered the 10 frames where the billboard had the largest bounding box for each billboard ID and each driver, resulting in 1,816 testing samples. We consider only billboard instances from the official test set of BillboardLamac. For aggregated classification a single class is provided for each billboard instance using all available testing data as in \cite{10308914}. To obtain the aggregated classes using our method, we simply perform voting using the predicted class probabilities, selecting the class with the highest aggregated probability. However, we note that the original features cannot be obtained from individual images and require multiple videos to extract, thus making per-detection classification impossible.

Table~\ref{tab:results} shows the results on the BillboardLamac dataset. We report the classification accuracy along with the macro and micro F1-scores. In terms of the aggregated accuracy the original features from \cite{10308914} provide the best results. Moreover, we show that including the detected bounding boxes and visual features together with the original features does not improve performance. We verified this conclusion for all combinations of features. We include only the best combination in the results for clarity. 

Our experiments show further that leaving out the original features significantly hurts the performance when considering the aggregate accuracy. This is especially the case when relying solely on the extracted DINOv2 features. In that case, in aggregate, the classifiers always select the same class (medium), resulting in very low F1-scores. This shows that relying solely on the extracted visual features is not sufficient when using the BillboardLamac dataset. We observed similar results when utilizing other strategies, such as training deep convolutional neural networks.

However, when combined with the known position of the bounding box our results show that even without using such features, we can achieve similar accuracy while making it possible to classify billboards by driver gaze duration from individual frames. The results show that using the DINOv2 features from the full image together with the position of the bounding box yields the best results, achieving 68.1\% accuracy, 47.4\% macro F1-score, and 64.6\% micro F1-score. When aggregated over multiple frames this method yields the aggregated accuracy of 69\% which is very similar to the result obtained by \cite{10308914}.


\subsection{Evaluation Using Google Street View}

Since our method is capable of classifying billboard gaze durations from single images, we perform additional evaluation. We use Google Street View to obtain images of the billboards from the test set of BillboardLamac. Using the functionality to view streets at different times we were able to retrieve images of 6 unique billboards from multiple views, yielding 30 images in total. See Figure~\ref{fig:gsv} for examples. We ran our best model using the bounding box position (\textit{B}) and the DINOv2 features from full images (\textit{I$_{full}$}), yielding an accuracy of 66.3\%, macro F1-score of 61.2\% and micro F1-score of 63.6\%. 

These results show that our method is able to generalize outside of videos captured using eye-tracking devices on drivers. These results also show potential for future applications of our method where the driver gaze duration can be estimated from single images of billboards such as those provided by Google Street View.

\section{\uppercase{Conclusion}}

This work introduced a two-stage pipeline for detecting roadside billboards from the driver’s perspective and classifying them according to their gaze duration. Our proposed approach aims to support studies on driver attention and advertisement impact within real-world driving environments. 

Compared to previous work, which requires the extraction of features from multiple videos with different drivers on the same route, our system can detect and classify billboards from single images. This significant benefit comes at a slight decrease of accuracy.


The current work has several limitations. For both training and evaluation, we rely on a dataset that contains large amounts of images, but only 145 instances of unique billboards. This places limits on the use of deep learning and visual features for this type of data due to the possibility of overfitting. Moreover, this limitation makes it difficult to properly evaluate the proposed methods as the variation of billboards in the test split is small and may not generalize well to other settings. As future work, it would be worthwhile to collect larger and more varied data, possibly employing some automation strategy that does not require significant manual labeling effort. Availability of such datasets may open new avenues for research in this area.


\label{sec:conclusion}

\section*{\uppercase{Acknowledgements}}

This work was funded by the EU NextGenerationEU through the Recovery and Resilience Plan for Slovakia under the project ''InnovAIte Slovakia, Illuminating Pathways for AI-Driven Breakthroughs" No.~09I02-03-V01-00029.

\bibliographystyle{apalike}
{\small
\bibliography{example}}

\end{document}